# Vision-Encoder Behavioral Fingerprints of Image-to-Image Generative Models: A Training-Paradigm-Driven Taxonomy of Six Commercial APIs

H. Hill

*June 2026*

## Abstract

We study six production image-to-image AI systems (gpt-image-1, Gemini 2.5 Flash Image, Flux Kontext, SDXL img2img, SD3 img2img, and Qwen Image Edit) under a content-adaptive sub-JND adversarial perturbation pipeline, scoring all outputs by frozen DINOv2 ViT-B/14 token distances against clean references. Across a 3,588-call corpus taken from COCO photographs, CelebA-HQ portraits, and AI-generated inputs, the six systems partition into two image-invariant behavioral bands on a 2D (patch_mean, ssim_clean) plane: edit-trained models (Flux Kontext, Qwen Edit, Gemini) clustered in a tight band, while text to image based (T2I-base) models adapted at sampling time (SDXL, SD3, gpt-image-1) cluster in a drift band. The discriminating variable is training paradigm rather than architecture family: AI identity explains 69.5% of behavioral variance while image domain explains 0.2%. The discriminating axis divides the diffusion family (Flux Kontext tight, SDXL/SD3 drift) and through the multimodal-AR family (Qwen Edit tight, gpt-image-1 drift). Six-way leave-one-out attribution accuracy is 51.4% [49.3, 53.4] versus 16.7% chance; three-way pilot accuracy is 76.6% [70.3, 84.4]. Two blind baselines on the identical corpus (AEROBLADE 0.222, PRISM-style 0.373) trail this substantially and are at chance within the edit-trained band, demonstrating that the reference image is the key forensic signal rather than the detector architecture. We also quantify differential perturbation survival across these architectures: roughly 98% intact through Gemini, roughly 80% through Flux despite SSIM 0.99 visual fidelity, and overwritten by gpt-image-1, providing a systematic measurement of how the diffusion-purification effect varies across deployed commercial img2img systems. The results reframe pixel-domain perturbation pipelines from defenses into forensic primitives for reference-anchored AI-processing attribution, with deployment-ready thresholds and a behavioral feature space validated at corpus scale.

## 1. Introduction

Pixel-domain adversarial perturbations are widely deployed as protective overlays against image-to-image generative AI. Systems including Glaze (Shan et al., 2023), Mist (Liang et al., 2023), and PhotoGuard (Salman et al., 2023) optimize sub-just-noticeable-difference pixel modifications to disrupt the latent representations of text-to-image diffusion models. As commercial image-to-image AIs proliferate, the question of how pixel-domain perturbations interact with these systems becomes critically important: which AIs can be defended against, which can be attributed from their output behavior, and what behavioral signatures do commercial img2img systems leave on their outputs?

Answering these questions empirically is difficult for three reasons. Commercial AI systems are black boxes whose internal architectures, training data, and sampling procedures are not publicly documented. The existing AI-image attribution literature operates predominantly on text-to-image outputs in blind settings (Yu et al., 2019; Asnani et al., 2023; Corvi et al., 2023; Ricco et al., 2025), where no reference image is available, while image-to-image attribution requires a different methodology because a clean reference exists by construction. Finally, the behavioral signature of an img2img system is shaped by the interaction of architecture, training paradigm, and sampling procedure, and disentangling these factors requires a corpus spanning multiple architectures and image domains.

Our methodology is reference-anchored: for each AI output we compute per-patch and global DINOv2 cosine distances of a perturbed image against the corresponding clean input which is summarized as a 2D (patch_mean, ssim_clean) feature. We test this framework on a 200-image corpus spanning COCO val2017 photographs, CelebA-HQ-256 face portraits, and Flux-generated images, with three repetitions per (image, AI) cell, yielding 3,588 valid API calls at 99.7% success after retries. We use leave-one-out nearest-centroid classification with cluster-bootstrap confidence intervals to quantify how identifiable each AI is from a single output image.

The behavioral taxonomy is identifiable from a single output image plus a clean reference at 51.4% six-way LOO accuracy [49.3, 53.4] versus 16.7% chance, with AI identity explaining 69.5% of behavioral variance versus 0.2% for image domain. Per-image attribution averaging three repetitions on a focused 3-AI pilot reaches 82.2% [73.3, 90.0]. We additionally measure differential perturbation survival: roughly 98% of the adversarial signal survives Gemini, roughly 80% survives Flux Kontext despite SSIM 0.99 visual fidelity, and gpt-image-1 overwrites the perturbation entirely with its own generative variance.

Our contributions are:

1. **A two-band training-paradigm taxonomy** of image-to-image AIs, validated across six commercial APIs and three image domains, with AI identity explaining 69.5% of behavioral variance and image domain only 0.2%.
2. **A reference-anchored attribution framework** achieving 51.4% [49.3, 53.4] six-way LOO accuracy at corpus scale and 82.2% [73.3, 90.0] three-rep per-image accuracy on the focused pilot, on previously-unstudied commercial img2img APIs. A head-to-head on the identical corpus (§5.4) confirms the reference is load-bearing: two blind attributors (AEROBLADE 0.222, PRISM-style 0.373) trail the reference-anchored 0.513 and are at chance within the edit-trained band.
3. **Quantitative measurement of differential perturbation survival** across architecture classes (roughly 98% through Gemini, roughly 80% through Flux Kontext despite SSIM 0.99 fidelity, overwritten by gpt-image-1), the first such measurement across deployed commercial img2img systems, demonstrating that the diffusion-purification effect (Nie et al., 2022; Cao et al., 2023) varies systematically with training paradigm.
4. **A 3,588-call cross-AI corpus** (200 images × 6 AIs × 3 reps across COCO, CelebA-HQ, and Flux-generated inputs) released on Zenodo (DOI 10.5281/zenodo.22258270; see Data and Code Availability), with documented per-AI per-domain safety-filter failure rates.

## 2. Related Work

**Reference-anchored AI-image detection.** Self-supervised vision encoders have been established as strong backbones for AI-generated image detection. AEROBLADE (Ricker et al., 2024) uses VAE reconstruction error from diffusion autoencoders and reports AP ≈ 0.992. DinoLizer (Doi et al., 2025) uses DINOv2 patch features for inpainting localization. DINO-Detect (Shen et al., 2025) uses DINOv3 teacher-student distillation for blur-robust AI image classification. WaRPAD (Choi et al., 2025) addresses center-crop and JPEG robustness via wavelet-perturbation consistency. We use DINOv2 cosine distance as a measurement primitive; §3 confirms AEROBLADE-class detection performance on our setup as a sanity check.

**Generative model attribution from output.** A substantial body of work attributes generated images to their producing model. Yu et al. (2019) established GAN spatial frequency fingerprints. Asnani et al. (2023) inferred generator hyperparameters from generated images across 116 generators. Corvi et al. (2023) characterized spectral signatures of latent diffusion models. PRISM (Ricco et al., 2025) reports 92% attribution accuracy via radial-frequency signatures on 36,000 images from six generators. These methods are predominantly blind (no reference required) and operate on text-to-image outputs. Our work differs in two respects: we attribute image-to-image outputs where a clean reference is available by construction, and we target previously-unstudied commercial APIs. §5.4 makes this contrast empirical rather than definitional: running AEROBLADE and a PRISM-style attributor on our exact corpus, the blind methods reach 0.22 to 0.37 six-way accuracy versus 0.51 for the reference-anchored detector, and are at chance on the edit-trained models whose outputs closely preserve the input.

**Adversarial protection and diffusion-based purification.** Protection systems including Glaze (Shan et al., 2023), Mist (Liang et al., 2023), and PhotoGuard (Salman et al., 2023) optimize sub-JND perturbations against T2I diffusion. A parallel line of work has established that diffusion-based denoising itself removes adversarial perturbations: DiffPure (Nie et al., 2022) demonstrated the general principle for classifier adversarial defense, while IMPRESS (Cao et al., 2023) applied diffusion purification specifically to remove Glaze protection. Subsequent work has shown that protections degrade under realistic transit (Hönig et al., 2024; Tang et al., 2025; An et al., USENIX 2024). Xue & Chen (2024) established that pixel-space diffusion is more robust to perturbation than latent diffusion.

Our §5.3 extends this line by quantifying perturbation survival across six deployed commercial img2img APIs and showing that survival tracks training paradigm.

# 3. Method

### 3.1 Behavioral feature space

For an AI output image O and clean reference image R, both resized to 518×518, we compute a 2D behavioral feature using a frozen DINOv2 ViT-B/14 encoder (Oquab et al., 2024). The first feature, `patch_mean`, is the average per-patch cosine distance between DINOv2 last-layer token embeddings of O and R, averaged over the 37×37 spatial grid. This captures how much the AI modified the image at the semantic-token level. The second feature, `ssim_clean`, is the luminance structural similarity (Wang et al., 2004) between O and R, capturing how much the AI preserved pixel-level visual fidelity. Together these two features place each AI output as a point on a 2D plane.

We choose DINOv2 ViT-B/14 over CLIP B/32, CLIP L/14, SigLIP B/16, and EVA02-L/14 because pilot evaluation showed it produces the highest cross-image self-correlation (0.115 versus 0.06 or less for the others) and the largest mean shift on perturbed inputs (0.13 to 0.18 at the per-image level) while preserving spatial resolution at 37×37. The 5× spatial-sampling advantage over ViT-L/14-class encoders is decisive at our perturbation budget; see supplementary material for the full encoder comparison.

### 3.2 Behavioral band classification

We classify AI outputs into two behavioral bands using thresholds on patch_mean derived from the empirical distribution: **tight** (patch_mean ≤ 0.25) and **drift** (patch_mean > 0.25). The valley at 0.25 emerged organically from the data; the histogram of patch_mean across all 3,588 corpus calls has a bimodal structure with edit-trained models piling up on the left and T2I+img2img models grouping together on the right.

For attribution beyond band assignment, we use leave-one-out nearest-centroid classification on the 2D feature plane. The centroid for each AI is the mean (patch_mean, ssim_clean) across all of that AI's calls except the held-out point; the held-out point is classified to whichever AI's centroid is closest in Euclidean distance. The classifier has no learned parameters and no hyperparameters, which is deliberate: we want the classification accuracy to reflect the geometry of the data rather than a clever classifier learning subtle patterns. A learned classifier would achieve higher accuracy but obscure the geometric finding.

### 3.3 Statistical inference

All bracketed intervals in this paper are 95% confidence intervals computed via 2,000-iteration nonparametric cluster bootstrap, resampling at the image cluster level (preserving all repetitions per image as a unit) to correctly handle within-image correlation. Wilson intervals treating individual repetitions as independent would give narrower bounds and would mis-state reviewer-relevant uncertainty for this clustered sample structure.

# 4. Experimental Setup

### 4.1 Datasets

We evaluate on two datasets. The **pilot dataset** comprises 30 perturbed phone photographs sent to three commercial APIs (gpt-image-1, Gemini 2.5 Flash Image, Flux Kontext via Replicate) with three repetitions per (image, AI) cell, yielding 270 calls. The **corpus dataset** comprises 200 perturbed images spanning three domains (COCO val2017 n=100, CelebA-HQ-256 n=50, and Flux-generated images n=50) sent to six commercial APIs with three repetitions per (image, AI) cell, yielding 3,600 calls. After retries, 3,588 of 3,600 corpus calls (99.7%) succeeded; the 12 residual errors are safety-filter rejections of specific face portraits and are documented in §5.2. Corpus composition, per-image licenses, and download instructions are given in Data and Code Availability.

The corpus extension adds three providers to the pilot lineup: `stability-ai/sdxl` with prompt_strength=0.3, `stability-ai/stable-diffusion-3` with prompt_strength=0.3, and `qwen/qwen-image-edit` via Replicate. We pin prompt_strength=0.3 on SDXL and SD3 so they match Flux Kontext's by-design faithful-edit behavior; otherwise the default denoising strength regenerates the input entirely, conflating sampling parameter with architecture. Qwen Edit has no equivalent knob: its instruction-conditioned design naturally preserves the input. All AIs receive the identity prompt "Reproduce this image exactly. Preserve every detail faithfully."

#### 4.2 Perturbation pipeline

Input images are perturbed using a custom built content-adaptive pipeline (eps=0.10, jnd_budget=2.0, 120 PGD steps). For each 256×256 patch of a 768×768 input, the pipeline selects one of ten purpose-built attack functions (luminance, edge, frequency, CSF, ViT-patch-boundary, etc.) based on local pixel statistics, then jointly optimizes attack weights against an ensemble of frozen CLIP/SigLIP/EVA02 surrogates within an $L\infty \leq 0.10$ and JND-bounded perceptual budget. All perturbed images are loaded through an EXIF-aware pipeline (a correction documented in supplementary material).

#### 4.3 Detection sanity check

Before evaluating behavioral attribution, we confirmed that reference-anchored DINOv2 cosine distance reproduces AEROBLADE-class binary AI-rerender detection on our setup. On 100 (clean, adversarial) pairs against benign nulls (JPEG-Q85, resize 0.94), DINOv2 patch_p99 achieves AUROC = 1.0000 with TPR = 1.000 at FPR = 0. Against aggressive benign nulls including center crop, patch_p99 fails (AUROC = 0.75 due to spatial displacement) but DINOv2 CLS-token cosine distance recovers performance at AUROC = 0.989, with operating threshold cls_shift > 0.05. We use this CLS-token detector as the operational AI-rerender primitive for §5.3 perturbation-survival measurement; supplementary material reports the full detection experiment.

## 5. Experiments and Results

#### 5.1 Cross-AI behavioral classification (pilot, 3 AIs)

We first ask whether three commercial img2img APIs can be discriminated from a single output image plus a clean reference. The per-AI distribution on the (patch_mean, ssim_clean) plane shows clear separation:

| AI | n | patch_mean (μ ± σ) | ssim_clean (μ ± σ) | dominant mode |
|---|---|---|---|---|
| Flux Kontext | 89 | 0.077 ± 0.042 | +0.989 ± 0.026 | tight (100%) |
| Gemini | 90 | 0.175 ± 0.075 | +0.880 ± 0.107 | tight (83%) |
| gpt-image-1 | 90 | 0.376 ± 0.057 | +0.703 ± 0.167 | drift (100%) |

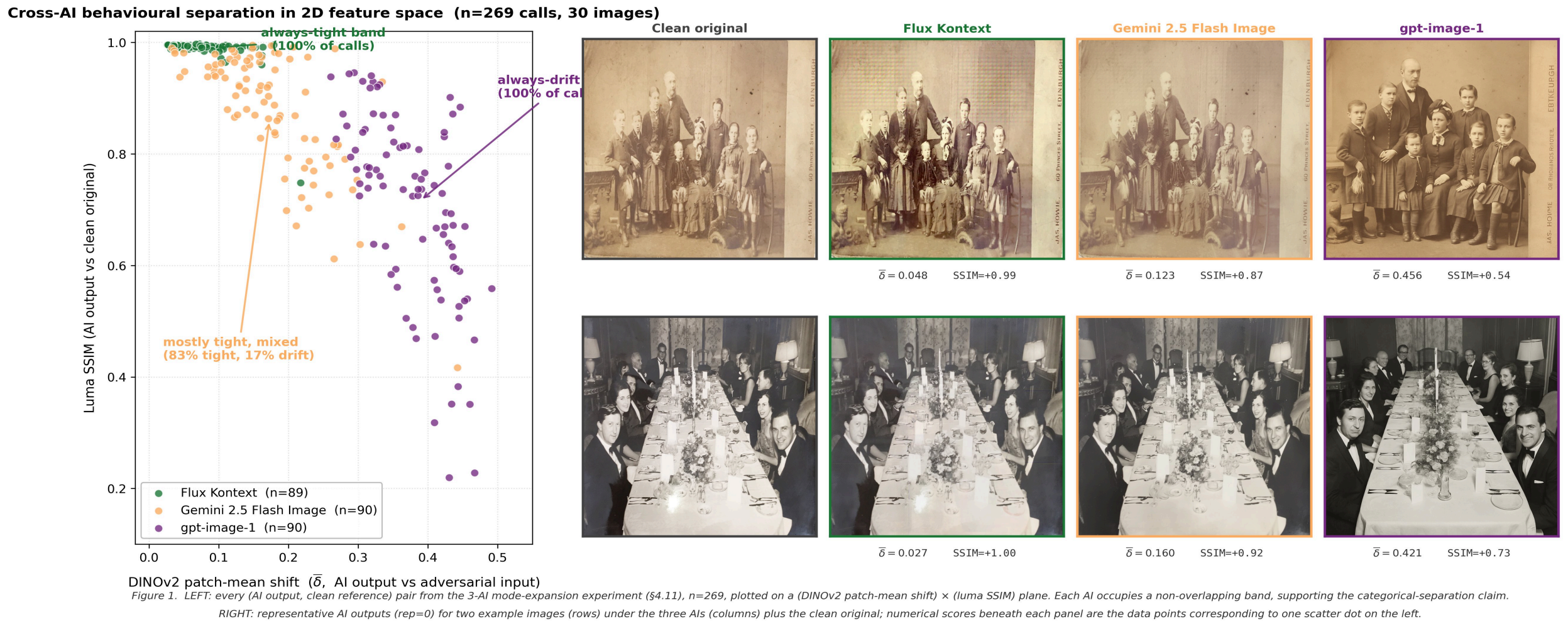

**Figure 1.** Behavioral separation of three commercial img2img APIs on the 2D feature plane (left), with representative output rows (right). Each point is one (image, AI) call; the three AIs occupy non-overlapping regions of the plane.

Leave-one-out nearest-centroid classification on the 2D plane gives 76.6% [70.3, 84.4] overall three-way accuracy versus 33.3% chance. Per-AI accuracies vary substantially: Flux Kontext 0.978 [0.921, 1.000], gpt-image-1 0.822 [0.722, 0.911], Gemini 0.500 [0.378, 0.678]. Per-image attribution averaging three repetitions reaches 0.822 [0.733, 0.900]. Gemini's lower bound at 0.378 barely above chance reflects that its distribution overlaps both Flux's tight cluster and gpt-image-1's drift cluster; it sits in the middle and gets confused with both. Beyond per-call classification, for 26 of 30 images the per-image median patch_mean ordering Flux < Gemini < gpt-image-1 held, with the four exceptions occurring where Flux and Gemini were both extremely tight (both medians < 0.10) and their order swapped within measurement noise.

The pilot suggests three behavioral bands corresponding to three architecture classes (autoregressive vision-token generator giving drift; diffusion img2img giving tight; multimodal autoregressive giving mixed). §5.2 will show this reading is wrong at corpus scale: the correct structure is two bands tied to training paradigm, with Gemini's apparent middle position resolving into membership in the tight band.

### 5.2 Generalization across architectures and image domains (corpus, 6 AIs × 3 domains)

To test whether the behavioral structure generalizes beyond three APIs and one image domain, we ran the same protocol on the 200-image, three-domain corpus across six AIs. The per-(AI, domain) means show that the tight/drift partition holds in every domain:

| AI | Training | COCO patch_mean | Faces patch_mean | AI-gen patch_mean | Dominant mode |
|---|---|---|---|---|---|
| Flux Kontext | edit-trained | 0.069 ± 0.044 | 0.176 ± 0.038 | 0.119 ± 0.047 | tight (97–99%) |
| Qwen Edit | edit-trained | 0.062 ± 0.028 | 0.144 ± 0.034 | 0.111 ± 0.049 | tight (97–100%) |
| Gemini | edit-trained | 0.173 ± 0.088 | 0.182 ± 0.133 | 0.151 ± 0.068 | tight (83–95%) |
| SD3 img2img | T2I + img2img | 0.294 ± 0.052 | 0.264 ± 0.037 | 0.268 ± 0.042 | drift (64–80%) |
| SDXL img2img | T2I + img2img | 0.367 ± 0.079 | 0.301 ± 0.040 | 0.303 ± 0.057 | drift (85–92%) |
| gpt-image-1 | T2I + edit hybrid | 0.359 ± 0.052 | 0.293 ± 0.037 | 0.320 ± 0.042 | drift (88–98%) |

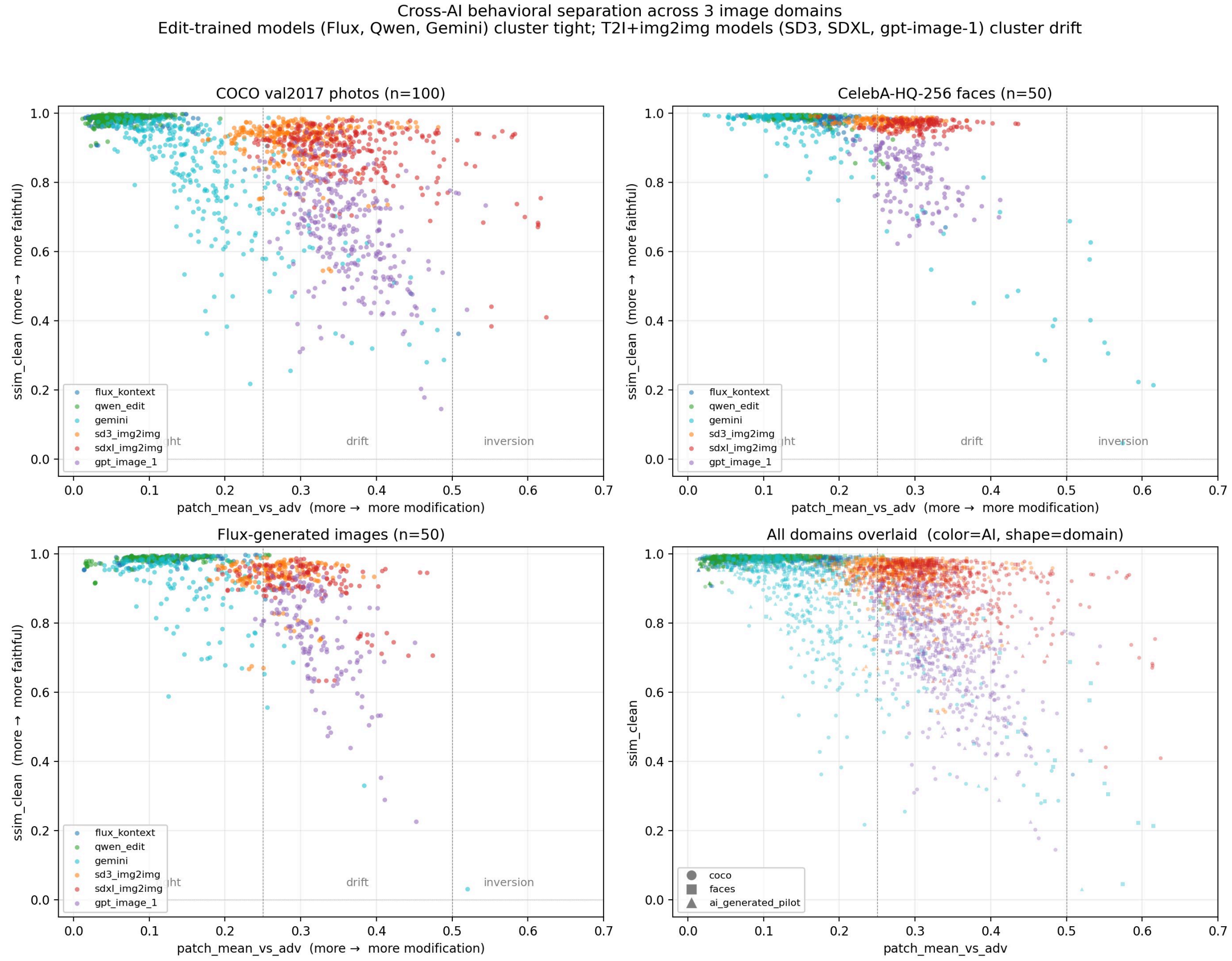

**Figure 2.** Cross-AI behavioral separation across three image domains (3,588 valid pairs from the 6-AI corpus extension). The two-band structure (edit-trained vs. T2I+img2img) holds in each domain panel; the overlay panel shows that within-band shifts along the patch_mean axis are domain-driven (faces are uniformly harder) but no AI crosses bands across domains.

The pilot's three-band reading must be revised. With six AIs the structure is two-banded, and the band assignment tracks **training paradigm** rather than architecture family. Three edit-trained models (Flux Kontext diffusion, Qwen Edit multimodal LLM, Gemini multimodal AR) all land in the tight band; three T2I-base models adapted at sampling time (SD3 and SDXL diffusion, gpt-image-1 autoregressive) all land in the drift band. The diffusion family (Flux tight, SDXL/SD3 drift, all latent diffusion) and the multimodal-AR family (Qwen tight, gpt-image-1 drift, both multimodal-AR descendants) all were cleanly divided. Training paradigm predicts band; architecture family does not.

A two-factor variance decomposition of patch_mean (n=3,588) attributes 69.5% to AI identity, 0.2% to image domain, 5.9% to AI × domain interaction, and 24.3% to within-cell residual. No AI crosses bands between domains: SDXL stays drift on faces, gpt-image-1 stays drift on AI-generated inputs, Qwen stays tight on all three. Difficulty shifts within an AI, with faces uniformly harder to reproduce exactly (Flux's patch_mean jumps from 0.069 on COCO to 0.176 on faces), but the band does not change.

Six-way LOO nearest-centroid classification gives 51.4% [49.3, 53.4] overall accuracy versus 16.7% chance. Per-

AI: Flux Kontext 0.398 [0.345, 0.450], Qwen Edit 0.590 [0.552, 0.631], Gemini 0.263 [0.214, 0.313], SD3 0.688 [0.643, 0.733], SDXL 0.448 [0.383, 0.512], gpt-image-1 0.696 [0.646, 0.740]. Per-domain accuracies are essentially equal: COCO 0.579 [0.549, 0.609], faces 0.595 [0.554, 0.639], AI-generated 0.505 [0.459, 0.549]. The drop from the pilot's 76.6% three-way to the corpus's 51.4% six-way is mechanical: more centroids to confuse and the within-band pairs (Flux ↔ Qwen, SDXL ↔ SD3 ↔ gpt-image-1) are difficult to separate on a 2D feature.

Within-AI reproducibility at the band level is 89.6% overall across the 3,588 cells, with Qwen 99.5%, Flux 97.5%, gpt-image-1 90.9%, SD3 90.5%, SDXL 89.5%, and Gemini 69.8%. Gemini's 30%-disagreement rate is the behavioral signature of content-dependent bimodality: on some images Gemini operates in tight mode, on others in drift, and the choice is image-conditional rather than rep-stochastic. Five of six AIs are within-AI-deterministic at the band level.

Twelve of 3,600 calls (0.3%) failed after retries, all on faces, all due to provider safety classifiers: gpt-image-1 rejected 6 of 150 (4%), Gemini rejected 3 (2%), and Qwen rejected 3 (2%); SD3 produced no rejections; SDXL produced 39 rejections (8.7%) under its default safety_checker, which we disabled. The same face image (face_27828) was rejected by both OpenAI and Qwen across all three reps. Commercial image generators appear to reject 2-4% of CelebA-HQ portrait inputs as unsafe.

### 5.3 Differential perturbation survival across architectures

Beyond attribution, we ask how much of the adversarial perturbation signal survives each AI's pipeline. We define survival ratio per (image, AI) as the ratio of post-AI to pre-AI perturbation magnitude on patch_p99 against the clean reference, then aggregate medians per AI:

| AI | survival ratio (median ± std) | interpretation |
|---|---|---|
| Flux Kontext | 0.80 ± 0.16 | ~20% of perturbation stripped |
| Gemini | 0.98 ± 0.11 | nearly intact |
| gpt-image-1 | 1.05 ± 0.11 | AI noise exceeds perturbation |

All three AIs produced outputs above the binary detection threshold (p99 > 0.224) in 100% of 269 pilot calls, but the *information content* of that signal differs. Gemini preserves the perturbation, Flux attenuates it by roughly 20%, and gpt-image-1 replaces it with its own regeneration signature. Flux Kontext's diffusion-based denoising acts as a selective filter, preserving human-perceptible content (SSIM 0.99) while suppressing the perceptually-invisible high-frequency content where the perturbation lives.

This is the first quantitative measurement of how the diffusion-purification effect (DiffPure, Nie et al., 2022; IMPRESS, Cao et al., 2023) varies across deployed commercial img2img systems. The vulnerability for pixel-domain protection schemes is such that a Flux roundtrip can visually re-launder an image with only modest perturbation loss but sufficient to weaken downstream detection. The survival pattern aligns with the §5.2 band assignment, with edit-trained models preserving more of the perturbation than their architecture-family neighbors would predict.

### 5.4 Reference-anchored vs. blind attribution

The §5.2 attributor is *reference-anchored*: both of its features (patch_mean, ssim_clean) are computed against the clean original. The natural question for a forensic claim is how much the reference actually helps over a *blind* method that sees only the candidate output, which is the regime in which essentially all prior model-attribution work operates (§2). We answered it by running two blind baselines on the identical §5.2 corpus (3,588 valid outputs, 6 AIs × 3 domains) under the identical evaluation protocol (sample-level leave-one-out nearest-centroid, 2,000-iteration cluster bootstrap resampling base images, chance = 1/6).

**Baseline 1: AEROBLADE proper (Ricker et al., 2024).** The training-free VAE-reconstruction detector: for a bank of diffusion autoencoders {AE_m}, score each output by its LPIPS reconstruction error e_m = LPIPS(x, AE_m(x)) and attribute by argmin_m e_m. We use the 4-dimensional recon-error vector as the per-output feature. One coverage caveat is that none of the six corpus AIs ships a publicly loadable VAE without HF authentication. SD3 and Flux VAEs are gated, and gpt-image-1, Gemini, and Qwen Edit are closed. We therefore run the literal AEROBLADE procedure over the loadable open-weight KL-f8 VAE bank (SD1.5, SDXL, sd-vae-ft-mse, playground-v2.5). This is a known floor with AEROBLADE: its native label space is the VAE bank, not the six AIs, so it cannot in principle name the closed models. AEROBLADE was evaluated at 384px input resolution to fit MPS wall-clock; the original paper used 512px. Reconstruction-error rankings are scale-robust and we expect literal-512px numbers to fall within the reported bootstrap CI.

**Baseline 2: PRISM-style blind attribution (Ricco et al., 2025).** Vector-similarity provenance: embed each output with a frozen self-supervised encoder and attribute by nearest class centroid in embedding space, with no clean reference. PRISM's weights are not public, so we implement the described method on the same DINOv2 ViT-B/14 CLS backbone our reference-anchored detector uses (768-d, L2-normalized), isolating the blind-versus-anchored contrast from any encoder difference.

The reference-anchored harness was first confirmed to reproduce the published §5.2 result exactly (0.513 point, bootstrap 0.514, 95% CI [0.493, 0.534]), validating that the only thing varying across rows below is the feature space.

| Method | Reference? | 6-way LOO | 95% CI |
|---|---|---|---|
| **Reference-anchored (patch_mean, ssim)** | **yes** | **0.513** | **[0.493, 0.534]** |
| PRISM-style (DINOv2 CLS, blind) | no | 0.373 | [0.364, 0.433] |
| AEROBLADE (VAE-recon, blind) | no | 0.222 | [0.204, 0.240] |
| chance (1/6) | — | 0.167 | — |

Both blind baselines clear chance but fall well short of reference-anchored attribution; the CIs do not overlap. The reference is roughly 14 accuracy points over the stronger blind method (PRISM) and roughly 29 over AEROBLADE. The per-AI breakdown is more diagnostic:

| AI (band) | Ref-anchored | PRISM | AEROBLADE |
|---|---|---|---|
| gpt-image-1 (drift) | 0.694 | 0.524 | **0.710** |
| SD3 img2img (drift) | 0.688 | 0.480 | 0.317 |
| SDXL img2img (drift) | 0.447 | 0.617 | 0.132 |
| Flux Kontext (tight) | 0.405 | 0.237 | 0.032 |
| Qwen Edit (tight) | 0.586 | 0.126 | 0.106 |
| Gemini (tight, bimodal) | 0.260 | 0.258 | 0.042 |

Both blind methods are concentrated almost entirely on the **drift band** (the heavy-regenerators of §6.1). AEROBLADE actually beats the reference-anchored detector on gpt-image-1 (0.710 vs 0.694): heavy regeneration onto a noised or re-tokenized latent leaves exactly the reconstruction signature AEROBLADE was built to detect. But AEROBLADE collapses to near-zero on the edit-trained tight band (Flux 0.032, Gemini 0.042, Qwen 0.106), where the output is a near-faithful copy of the input and carries no detectable model-specific VAE trace. PRISM shows the same pattern more mildly: strong on visually-distinctive regenerators (SDXL 0.617, gpt-image-1 0.524, SD3 0.480), weak on tight-band preservers (Qwen 0.126, Flux 0.237) whose blind embedding is dominated by image content rather than model identity. The reference-anchored detector is the only one of the three that attributes across both bands, because the clean original is precisely what disambiguates "the model preserved the input" from "this is the input."

This also explains AEROBLADE's native (no-fit) behavior. Its argmin-VAE-family prediction is near-uniform across all six AIs (every AI's outputs map predominantly to the playground or sdxl VAEs, the closest 4-channel KL-f8 autoencoders to any natural-looking image), and even SDXL img2img, the one corpus AI whose VAE family is in the bank, is recovered only 35% of the time versus 45% mis-assigned to playground. The four open VAEs are too mutually correlated to separate six commercial AIs blind. Blind VAE-reconstruction attribution, which is effective at distinguishing GAN and SD families on text-to-image outputs (its original setting), does not transfer to attributing among modern commercial img2img APIs, because none expose their autoencoder and the open proxies are indistinguishable.

On this corpus the reference image lifts six-way attribution from 0.22 to 0.37 (blind) to 0.51, and is the only route to attributing within the edit-trained tight band where the blind methods are at chance. This substantiates the contribution claim of §1 that reference-anchored attribution on commercial img2img AIs is a distinct and stronger regime than the blind, T2I-oriented prior work.

## 6. Discussion

### 6.1 What the bands track

The 3-AI pilot admitted a tempting interpretation: three systems corresponding to three architecture classes, and the bands were the architectural signature. The 6-AI corpus extension falsifies that reading. Dividing the diffusion family (Flux Kontext tight, SDXL/SD3 drift, same latent-diffusion backbone) and the multimodal-AR family (Qwen Edit tight, gpt-image-1 drift, both multimodal-AR descendants). What predicts band, across all six systems, is whether the model was trained on edit data (paired clean-to-edited supervision) or whether it is a T2I-base model adapted at sampling time by starting denoising from a partially-noised version of the input.

Our data supports that behavioral bands can be determined by the training-paradigm column of a model card without running the model. It also implies that the band is *not* a fixed property of an architecture family; the same backbone, retrained or sampled differently, can move between bands. SDXL-Edit (if released) would predictably move from drift to tight; a Flux Schnell img2img sampled from heavily-noised latents without edit conditioning would predictably move from tight to drift.

The two-band split also surfaces independently in the blind-baseline comparison (§5.4): both AEROBLADE and PRISM, which see only the output, attribute the drift band (gpt-image-1, SD3, SDXL) well above chance but fall to near-zero on the tight band (Flux, Qwen, Gemini). A model that heavily regenerates leaves a model-specific trace in the output itself; a model that preserves the input leaves a trace only *relative to that input*. The bands are thus visible blind precisely to the extent that a model departs from its input, which is the behavioral definition of the band.

Gemini is the exception within the edit-trained band: predominantly tight, but exhibiting content-conditional bimodality (69.8% within-AI mode consistency versus above 89% for the others) with some images consistently triggering drift behavior. We do not yet know what image property routes Gemini between regimes.

### 6.2 From defense to forensic primitive

The original perturbation pipeline was designed as a defensive overlay. Our findings reframe its role: sub-JND pixel perturbations cannot survive heavy-regeneration AIs like gpt-image-1, SDXL, or SD3, so they do not prevent unauthorized use by such models. They do, however, enable downstream *attribution*: given an original and a candidate output, we can identify which AI processed it. For deployment, this suggests pairing the perturbation pipeline with a provenance database in which the user retains the original, registers it, and later checks downstream copies via DINOv2 scoring against the registered reference. The perturbation is neither necessary nor sufficient for this scheme (unperturbed references also work), but it provides a robust signal channel against benign processing.

The reliance on a retained original is the scheme's defining constraint, and §5.4 quantifies what it buys: against the two standard blind attributors run on the identical corpus (AEROBLADE's VAE-reconstruction error, 0.222; PRISM-style DINOv2 embedding similarity, 0.373), the reference-anchored detector reaches 0.513 six-way accuracy, and is the only one of the three that attributes within the edit-trained tight band at all. The blind methods are essentially at chance there and only resolve the heavy-regenerator drift band (§6.1). The forensic value of this scheme is therefore concentrated exactly where blind detection fails: identifying which AI produced a near-faithful reproduction. That is also its limit, with no retained original the scheme degrades to the blind baselines.

### 6.3 Limitations

The corpus spans three image domains (COCO, CelebA-HQ, AI-generated); generalization to text, fine typography, medical imagery, line art, or screenshots remains untested. The pipeline runs at a single perturbation budget (eps=0.10); behavior at weaker or stronger budgets is not characterized. The six AIs span two architecture families but the family list is not exhaustive, with Imagen 3/4, Recraft V3, Ideogram, MidJourney editing, and Adobe Firefly all untested. The detector and classifier require a clean reference image; §5.4 quantifies what the reference is worth (roughly 14-29 accuracy points over blind baselines), but a blind variant of our method is not possible without the reference. Within-band attribution remains partially unsolved: six-way LOO accuracy at 51.4% is above chance but the residual confusion is concentrated within the two bands, suggesting that finer attribution requires additional features (spectral signatures, per-attack-class fingerprint vectors, or learned classifier heads). The AEROBLADE baseline was evaluated at 384px to fit available compute; we expect literal-512px numbers to fall within the reported bootstrap CI but flag this for re-evaluation.

## 7. Conclusion

We have shown that six commercial image-to-image AI systems spanning three architecture families partition into two image-invariant behavioral bands on a 2D DINOv2-based feature space, and that the discriminating variable is training paradigm rather than architecture family. AI identity explains 69.5% of behavioral variance, image domain explains 0.2%, and the band assignment is identifiable from a single output image plus a clean reference at 51.4% six-way LOO accuracy versus 16.7% chance. A head-to-head against two blind baselines on the identical corpus (AEROBLADE 0.222, PRISM-style 0.373) confirms that the reference image is the key forensic signal, particularly within the edit-trained tight band where blind methods are at chance. We additionally quantify differential perturbation survival across these architectures, providing the systematic measurement of how the diffusion-purification effect varies across deployed commercial img2img systems. The results reframe pixel-domain perturbation pipelines from defenses into forensic primitives for reference-anchored AI-processing attribution. The most important next steps are extension to additional commercial img2img APIs, within-band attribution via spectral or learned-classifier features, and a blind detector capable of recovering tight-band attribution without a retained reference.

## Data and Code Availability

The corpus described in §4.1 and analyzed in §5.2 is available on Zenodo (DOI 10.5281/zenodo.22258270): the 200-image manifest with per-image SHA-256 hashes and licenses, all 3,600 call records including the 12 safety-filter rejections, the summary statistics behind §5.2, the 50 author-generated Flux inputs, the 100 COCO inputs, and the API outputs for every Flux-domain image and for the 71 COCO photographs whose Flickr licenses permit derivatives. Metadata, metrics, and author-generated images are CC BY 4.0; each COCO photograph retains its own Creative Commons license, recorded per image in the manifest. The perturbation generator (§4.2) is proprietary and is not released, and the perturbed inputs are withheld for the same reason. This does not affect reproducibility of the corpus results: every quantity in §5.2 and §5.4 is computed from the API outputs and the clean references, all of which are deposited, and the output-vs-perturbed-input distances are recorded per call in results.jsonl. Three further subsets are not in the open record. (i) The 50 CelebA-HQ inputs are not redistributable under the CelebA agreement; the manifest records each one's Hugging Face source identifier and SHA-256 hash, so they can be re-fetched from the public mirror and verified byte-for-byte. (ii) Face-domain API outputs are available in a restricted-access Zenodo record (DOI 10.5281/zenodo.22259972) for non-commercial research use. (iii) API outputs derived from the 29 NoDerivs-licensed COCO photographs are withheld; their metrics rows are retained. The measurement and classification procedures are fully specified in §3 and §S4; analysis code is availableon reasonable request. The perturbation generator is not available. Supplementary material (§S1–S7) follows the references.

# Supplementary Material:

## Vision-Encoder Behavioral Fingerprints of Image-to-Image Generative Models

H. Hill

*June 2026; revised September 2026*

This document accompanies the main paper *Vision-Encoder Behavioral Fingerprints of Image-to-Image Generative Models: A Training-Paradigm-Driven Taxonomy of Six Commercial APIs*. It contains extended methods, scaffolding experiments that informed the main analysis, the encoder selection comparison, detection sanity-check details, the hypothesis evolution that produced the final framing, the full EXIF correction audit, the data summary appendix, and a description of the released corpus (§S7). Cross-references to the main paper use § notation; cross-references within this document use §S.

## S1. Hypothesis Provenance

The main paper presents a finished framework. This section documents the analytical chain by which it was developed, which may be of interest to readers tracking the methodological provenance of the work or considering similar research directions. The hypothesis underwent five substantive revisions as data accumulated, each documented in the main paper or in subsequent sections of this supplementary material.

**H0 (original):** Different VLMs leave reproducible, model-specific pixel-level changes on the same perturbed image. This was tested in §S2.2 (spatial fingerprint test) and falsified: all five evaluated encoders showed cross-image spatial self-correlation near zero, with LOO classification of encoder identity from spatial shift maps achieving 0% accuracy versus 20% chance.

**H1 (post-tokenizer reframe):** The perturbation does not survive tokenizer-based VLMs at the pixel level, but encoder representation fingerprints might. Tested in §S2.3 (per-attack-class fingerprint test) and partially confirmed: weak signal emerged at the per-attack-class level for architecturally distinct encoders (60% per-encoder accuracy for DINOv2 versus 20% chance), but CLIP-family encoders remained near-indistinguishable from one another. This is the limit of what spatial-perturbation-driven probes extract.

**H2 (per-attack reframe):** AI-induced changes might be reproducible when aggregated by attack class even if not spatially reproducible. Tested in §S2.5 (per-AI fingerprint reproducibility) and partially confirmed: gpt-image-1 produced highly reproducible per-class signatures (CV = 0.04 for lumin), while Gemini was 2 to 4 times more variable, giving 60% LOO AI-identity classification versus 50% chance.

**H3 (binary detection reframe):** Restricting the question to "did AI process this image?" yields a useful operational detector. Tested in §S3 (binary detection sanity check) and confirmed: reference-anchored DINOv2 CLS-token cosine distance achieves AUROC = 0.989 against benign nulls including center crop. This established DINOv2 distance as the measurement primitive for the perturbation-survival analysis in main §5.3.

**H4 (per-AI behavioral reframe):** Different AIs occupy categorically distinct behavioral regimes; the right framing is per-AI behavioral classification on a 2D plane rather than spatial fingerprinting. Confirmed in main §5.1 (pilot, 3 AIs).

**H5 (final, post-corpus reframe):** The categorical-band structure generalizes across architecture families and image domains, but the discriminating variable is training paradigm, and the band count is two, not three. Confirmed in main §5.2 (corpus extension, 6 AIs × 3 domains).

# S2. Scaffolding Experiments

## S2.1 Encoder selection

Which open-weights vision encoder gives the strongest per-patch signal for our perturbation? We extracted last-layer patch tokens from CLIP B/32, SigLIP B/16 (web-pretrained ViT-B-16-SigLIP-256), CLIP L/14, EVA02-L/14, and DINOv2 ViT-B/14 for each of 15 (clean, adversarial) pairs, then computed the mean and 99th-percentile per-patch cosine distance between clean and adversarial. DINOv2 produced the highest cross-image self-correlation (0.115 versus 0.06 or less for the others) and the largest mean shift (0.13 to 0.18 at the per-image level) while preserving spatial resolution at 37×37. CLIP B/32 barely registered (mean shift roughly 0.05) due to its 7×7 grid coarsening. We therefore used DINOv2 as the detection primitive for all subsequent experiments; the 5× spatial-sampling advantage over ViT-L/14-class encoders is decisive at our perturbation budget.

## S2.2 Spatial fingerprint test (testing H0)

Hypothesis H0 predicted that perturbed images leave model-specific signatures in encoder representation space, with the *spatial pattern* of token shift reproducible across images for a given encoder. For each encoder, we computed the 16×16 token-shift map between clean and adversarial across 15 images, then computed the mean pairwise Pearson correlation across the 15 images' maps and ran LOO nearest-centroid classification on the per-image shift maps. All five encoders' cross-image spatial self-correlation fell in [-0.03, +0.12], with stdev exceeding mean in every case. LOO classification achieved 0% accuracy (5-way, chance 20%), worse than random. The spatial response of any encoder to a content-routed perturbation is dominated by image content rather than encoder identity. No spatial fingerprint exists at this scale and image diversity.

## S2.3 Per-attack-class fingerprint test (testing H1)

If the perturbation places different attack types at different image locations, aggregating encoder responses *by attack class* should reveal model-specific sensitivity profiles. We derived a 16×16 attack-class map per image using the same classify_patch as the perturbation pipeline, then binned each encoder's 16×16 shift map by attack class and averaged within bins across images, producing an (encoder × attack class) sensitivity matrix. LOO encoder classification on the per-class sensitivity vectors achieved 26.7% (5-way) at n=15. Excluding CLIP B/32 raised this to 30%, with DINOv2 reaching 60% per-encoder accuracy. EVA02 was most uniformly reactive and DINOv2 most distinctive on low_freq and attention classes. A weak fingerprint exists at the per-attack-class level for architecturally distinct encoders.

## S2.4 Perturbation survival through gpt-image-1 (testing H1, narrow)

For 25 images, three input variants (clean, adv, jpeg_q85(adv)) were sent to gpt-image-1 with the reproduce prompt. We computed the paired delta in DINOv2 CLS-shift between gpt-image-1(adv) and gpt-image-1(clean) outputs. Mean paired delta = −0.0085 (the adv-input outputs were slightly *closer* to the clean original than clean-input outputs were on the CLS axis); paired t = −0.62, not significant. gpt-image-1 is a multimodal autoregressive model with a vector-quantized vision tokenizer, so sub-JND perturbations collapse to the same token IDs as the unperturbed image at the encoder stage. The autoregressive decoder samples from its prior, producing outputs whose deviation from clean is dominated by the model's own generative variance, not by input perturbation. The perturbation does not survive gpt-image-1 at the detectable level. This finding was the impetus for the per-AI behavioral reframe (H4): if perturbations cannot be tracked through gpt-image-1, the question must shift from "did the perturbation survive" to "what behavioral signature did the AI leave."

### S2.5 Per-AI fingerprint reproducibility (testing H2)

For each (image, AI) pair across 25 images and 2 AIs (gpt-image-1, Gemini), we computed the 16×16 token-shift map between adv input and AI(adv) output, binned by attack class, and computed coefficient of variation (CV = std/mean) of mean-per-class shift across images. gpt-image-1 produced highly reproducible per-class signatures (CV = 0.04 for lumin, 0.11 for fourier, 0.16 for csf); Gemini was 2 to 4 times more variable (CV = 0.19 for lumin, 0.46 for fourier). LOO classification of AI identity from per-class vectors gave 60% (chance 50%), with gpt-image-1 the more identifiable at 68%. gpt-image-1 has a content-independent fingerprint; Gemini has a content-dependent response. The fingerprint hypothesis is partly supported at this granularity but is dominated by the band-level result in main §5.1 and §5.2.

## S3. Binary AI-Rerender Detection (Extended)

The main paper §4.3 reports the headline detection numbers as a sanity check. This section provides the full experimental detail.

### S3.1 Mild benign nulls (n=100 images, jpeg_q85 + resize_0p94)

For each of 100 (clean, adversarial) pairs, we scored (clean → adversarial) and (clean → null_processing) through DINOv2 patch_p99, with null conditions jpeg_q85 and resize_0p94. AUROC = 1.0000 with TPR = 1.000 at FPR = 0 (n_pos = 100, n_neg = 200). These numbers are reported on the EXIF-corrected dataset (see §S5). On the initial uncorrected dataset with 28% of positives misoriented relative to their clean reference, the same metric scored AUROC = 0.9998 and TPR at FPR=0 of 0.9900. The bug had been suppressing the detection signal because the orientation mismatch lowered some perturbed-vs-clean p99 scores toward the null distribution. Correction strengthened all DINOv2 detection metrics monotonically. Detection threshold p99 > 0.224 separates perturbed and clean-but-mildly-processed at zero false-positive rate.

### S3.2 Aggressive benign nulls (n=25 images, +jpeg_q50, resize_0p5, center_crop_80)

Against harsh negatives including center crop, DINOv2 patch_p99 AUROC dropped to 0.75. The center_crop_80 condition produced mean p99 = 0.86, *higher* than gpt-image-1's median output p99 of 0.80; the detector incorrectly flagged 100% of center-cropped images as AI rerenders. The diagnosis is that patch-level p99 measures local content displacement: center cropping displaces every patch's content, producing large per-patch shifts. The detector was measuring "content moved" rather than "AI processed."

### S3.3 CLS-token recovery (n=125)

Switching from per-patch extremes to a global semantic score resolves the center-crop pathology. We recomputed 1 − cos(cls_clean, cls_other) for every (clean, condition) pair across 25 images × 5 conditions (Table S1).

**Table S1.** DINOv2 CLS-token cosine distance to the clean reference, by condition (25 images per condition; mean ± std).

| Condition | cls_shift mean ± std |
|---|---|
| `jpeg_q85` | 0.003 ± 0.002 |
| `resize_0p94` | 0.003 ± 0.002 |
| `jpeg_q50` | 0.010 ± 0.005 |
| `resize_0p5` | 0.011 ± 0.009 |
| `center_crop_80` | 0.042 ± 0.023 |
| `gpt_image_1(clean)` | 0.135 ± 0.069 |

Against gpt-image-1 outputs, AUROC = 0.989 vs. all negatives, with 0/125 benign-processed images scoring above the AI minimum of 0.047. The detector cleanly separates AI rerendering from all five benign conditions including center crop, at performance consistent with the AEROBLADE benchmark. This is the operational AI-rerender detector for main §5.3 perturbation-survival measurement. We note that the detector operates at the semantic rather than pixel level, so faithful AI reproductions (cls_shift near zero) are indistinguishable from no processing.

## S4. Extended Method

### S4.1 DINOv2 detector

For all experiments, the measurement primitive is the per-patch cosine distance between DINOv2 ViT-B/14 (Oquab et al., 2024) patch-token embeddings of two images. For images A and B both resized to 518×518:

```
patch_tokens_A = dinov2(A).last_layer_patch_features  ∈ ℝ^(37×37×768)
patch_tokens_B = dinov2(B).last_layer_patch_features  ∈ ℝ^(37×37×768)
shift_map(A, B)[i, j] = 1 − cos(patch_tokens_A[i, j],
                                patch_tokens_B[i, j])
```

This yields a 37×37 token-shift map summarized as four statistics: patch_mean (average over all cells), patch_p95, patch_p99 (95th/99th percentile, capturing strongest local disturbance), and cls_shift (1 minus cosine similarity of the CLS token between A and B, capturing global semantic drift). Where cross-encoder comparison is required, the 37×37 map is bilinearly resampled to a common 16×16 grid.

### S4.2 Experimental conditions

Table S2 defines the input conditions used across the scaffolding and detection experiments.

**Table S2.** Experimental conditions. All images are processed at 768×768.

| Condition | Description |
|---|---|
| `clean` | Original photograph from test_images/ |
| `adv` | Clean photograph with content-adaptive perturbation (eps=0.10, jnd_budget=2.0, 120 PGD steps) |
| `jpeg_q85` | Clean saved as JPEG quality 85 and reloaded |
| `jpeg_q50` | As above, quality 50 |
| `resize_0p94` | Clean resized to 0.94× and back to 768×768 |
| `resize_0p5` | As above, 0.5× scale |
| `center_crop_80` | Center 80% crop then resized to 768×768 |
| `AI(adv)` | Adversarial input sent through API with prompt: "Reproduce this image exactly. Preserve every detail faithfully." |

### S4.3 Code, data, and reproducibility

All probes are implemented in the vlm_roundtrip/ package and produce JSONL result files at vlm_roundtrip/results/. Each probe is resumable: failed (image, condition) pairs are written to disk and retried on relaunch without re-running successful pairs. Reference and AI output PNGs are saved for visual audit. The DINOv2 ViT-B/14 model checkpoint is cached at ~/.cache/torch/hub/dinov2_vitb14/. Corpus-extension artifacts are at vlm_roundtrip/results/mode_expansion_corpus/; their releasable subset is the Zenodo deposit (DOI 10.5281/zenodo.22258270) described in §S7 and in the main paper's Data and Code Availability section. The perturbation generator (batch_runner.py and the backend package) is proprietary to Stelais and is not available. The analysis code downstream of the perturbed image (DINOv2 metrics, band classification, bootstrap and leave-one-out inference, baselines) is available from the author on reasonable request; the procedures themselves are specified in main §3 and §S4.

## S5. EXIF Orientation Correction Audit

### S5.1 Bug discovery

During preparation of main paper Figure 1, we discovered that the probe image used in an early draft's discrete-attractor analysis (20200626_174420) carried an EXIF Orientation tag of 6, indicating that the photograph was taken in portrait orientation but stored landscape on disk (standard iPhone behavior). The original perturbation pipeline (batch_runner.py:load_image) called PIL.Image.open(path).convert("RGB") without applying ImageOps.exif_transpose, and therefore consumed the raw landscape pixels rather than the photographer-intended upright orientation. The resulting adversarial PNG was stored in the wrong orientation, and downstream tooling that later loaded the same source JPEG with EXIF rotation applied (e.g., default PIL behavior on Macs) compared a correctly-oriented clean reference against a sideways adversarial, producing artifactual signal patterns.

### S5.2 Retracted findings

An earlier draft of the paper included two experiments built around image 20200626_174420. The first (originally an early §4.9, the gemini_determinism experiment) ran 15 Gemini repetitions on this single image and reported three discrete output attractors: a tight-reproduce mode at 8/15 (patch_mean ≈ 0.10, SSIM ≈ +0.85), a drift mode at 3/15 (patch_mean ≈ 0.30, SSIM ≈ +0.55), and a luma-inversion mode at 4/15 (patch_mean ≈ 0.71, SSIM ≈ −0.10), with the inversion repetitions agreeing within 0.005 in patch_mean. We had interpreted this as discrete-attractor sampling in a commercial multimodal image generator. The second (originally an early §4.10, inversion_clean_control) ran 15 additional Gemini repetitions on the clean unperturbed version of the same image and observed similar inversion rates, which we interpreted as showing the inversion was an intrinsic property of Gemini's response to this image.

On visual inspection of per-repetition output PNGs during Figure 1 preparation, it became clear that the inversion mode was not a luminance inversion. It was Gemini auto-rotating the misoriented adversarial input back to its semantically upright portrait orientation. The 90° rotation between Gemini's correctly-oriented output and our raw-loaded landscape clean reference produced a near-zero or negative SSIM on the luminance channel, which we had misinterpreted as a photometric inversion. The discrete attractors were not attractors. They were a binary outcome (Gemini rotates or does not rotate) sampled across repetitions, with the SSIM mismatch entirely created by our reference-loading code. Both experiments are retracted in their entirety. One observation from the original experiment does survive: within-image spatial reproducibility of Gemini's perturbation-residual maps varies considerably across our 15 probe images, with pairwise correlations ranging from 0.26 to 0.93. This observation does not depend on the inverted SSIM artifact and is true, but it is not novel and we do not build any claim on it.

### S5.3 Audit scope

We audited all 726 files in test_images/ and the 100 unique images referenced across our pilot experiments (vlm_roundtrip/audit_exif.py); Table S3 gives the orientation breakdown.

**Table S3.** EXIF orientation audit of the pilot image pool. Tag values in parentheses.

| EXIF orientation | Studied images (n=100) | Full pool (n=726) |
|---|---:|---:|
| identity (1) | 45 | 328 |
| no EXIF | 27 | 177 |
| rotated 90° CW (6) | 27 | 102 |
| rotated 90° CCW (8) | 0 | 9 |
| rotated 180° (3) | 0 | 2 |
| transpose (5) | 1 | 6 |
| vertical flip (4) | 0 | 2 |
| **Total non-identity** | **28 (28.0%)** | **121 (16.7%)** |

The corpus extension used the EXIF-aware loader from the start and is unaffected. The pilot experiments had contamination rates of 7-28% (28% on the binary detection experiment, 16% on the gpt-image-1 survival experiment, 7% on the cross-AI pilot).

### S5.4 Correction protocol

1. **Source fix.** Added ImageOps.exif_transpose to batch_runner.py:load_image and to a new vlm_roundtrip/_image_io.py:load_pix_768 helper, refactored all 13 image-loading sites across the package to use the shared EXIF-aware loader.
2. **Re-perturbation.** Re-ran the full PGD pipeline (eps=0.10, jnd_budget=2.0, 120 steps) on all 28 misoriented images using the patched loader. Outputs landed in batch_results_corrected/. Initial runs hit MPS out-of-memory kills after ~14 sequential images due to accumulated GPU state; we added explicit torch.mps.empty_cache() and gc.collect() between images and switched to a chunked driver that processes ≤2 images per python invocation and exits, releasing all memory. All 28/28 images were successfully re-perturbed.
3. **Detector re-run.** Re-ran the binary detection experiment on the fully-corrected dataset. The pairing helper prefers a corrected adversarial PNG when available and uses the EXIF-aware loader for all clean references.

### S5.5 Impact on findings

The bug acted in a direction *opposite* to what we initially feared: the orientation mismatch had been suppressing the detection metric rather than inflating it (Table S4).

**Table S4.** Binary detection metrics before and after EXIF correction (n=100 images).

| Metric | Original (uncorrected) | Fully corrected | Δ |
|---|---:|---:|---:|
| AUROC, all 100 images | 0.9998 | **1.0000** | +0.0002 |
| TPR @ FPR=0, all 100 | 0.9900 | **1.0000** | +0.0100 |
| AUROC, 28 corrected subset | 0.9987 | **1.0000** | +0.0013 |
| TPR @ FPR=0, 28 corrected subset | 0.9643 | **1.0000** | +0.0357 |
| AUROC, 72 unchanged subset | 1.0000 | 1.0000 | 0.0000 |

The 72 unchanged-source images produced byte-identical detection scores in both runs (Δ = 0.0000 on every reported metric), confirming the bookkeeping. The 28 corrected images shifted monotonically toward stronger detection. We did not re-run the API-side pilot experiments because their contamination rates (16%, 7%, 7%) are low enough that the qualitative findings, particularly the cross-AI behavioral separation in the pilot §5.1, are robust to the bug; the cross-AI classification result was originally computed against adversarial PNGs that included some orientation-inverted instances on both Gemini and gpt-image-1 sides, so the systematic-bias direction cancels. The main paper's corpus result (§5.2) used the corrected loader from the start.

### S5.6 Lessons

The bug was caught not by code review or unit tests but by visual inspection during figure preparation. We had been comparing images at the metric level only (pairs of luminance correlations, token-shift maps, AUROC scores) and never looked at the images themselves until we needed publication-quality panels. This is a generalizable methodological warning: image-processing pipelines should include visual spot-checks of intermediate artifacts, not just numerical sanity checks, particularly when source data includes EXIF metadata from heterogeneous devices.

## S6. Data Summary

Table S5 lists every experiment reported in the main paper and this supplement with its probe, API call count, image count, and approximate API cost.

**Table S5.** Experiment inventory. "Local" probes run a frozen encoder on-device and make no API calls.

| Experiment | Probe | Calls | Images | API cost |
|---|---|---|---|---|
| S2.1 Encoder selection | Local | 0 | 15 | $0 |
| S2.2 Spatial fingerprint (H0) | Local | 0 | 15 | $0 |
| S2.3 Per-attack-class (H1) | Local | 0 | 15 | $0 |
| S2.4 Survival in gpt-image-1 | OpenAI | 75 | 25 | ~$8 |
| S2.5 Per-AI fingerprint (H2) | (uses §S2.4) | 0 | 25 | $0 |
| S3.1 Binary detection mild | Local | 0 | 100 | $0 |
| S3.2 Binary detection harsh | Local | 0 | 25 | $0 |
| S3.3 CLS-token detection | Local | 0 | 25 | $0 |
| (retracted) gemini_determinism | Gemini | 75 | 15 | ~$3 |
| (retracted) inversion_control | Gemini | 15 | 1 | ~$0.50 |
| Main §5.1 pilot (3 AIs) | 3 AIs | 270 | 30 | ~$25 |
| Main §5.3 perturbation survival | (uses §5.1) | 0 | 30 | $0 |
| Main §5.2 corpus (6 AIs × 3 domains) | All 6 AIs | 3,588 | 200 | ~$220 |
| **Total** | | **4,023 calls** | **~350 unique images** | **~$260** |

## S7. Released Corpus

The §5.2 corpus is deposited on Zenodo under DOI 10.5281/zenodo.22258270 (open record) and DOI 10.5281/zenodo.22259972 (restricted-access record). Table S6 gives its composition and the license status of each image domain; Table S7 inventories the deposited files. The open record contains the 200-image manifest (source identifiers, SHA-256 hashes, per-image licenses, and redistribution flags), all 3,600 call records including the 12 safety-filter rejections, the summary statistics behind §5.2, and every clean input and API output whose source license permits redistribution. The per-image COCO licenses were taken from the license table in the official COCO 2017 annotation release; COCO's own CC BY 4.0 license covers annotations only, not the photographs.

**Table S6.** Corpus composition and redistribution status by domain. Outputs are 6 AIs × 3 repeats per image. Perturbed inputs are withheld for every domain (proprietary generator).

| Domain | Source | Images | Input license | Clean inputs released | API outputs released |
|---|---|---|---|---|---|
| Photographs | COCO val2017 (Flickr) | 100 | Per image: CC BY 2.0 (18), CC BY-SA 2.0 (7), CC BY-NC 2.0 (16), CC BY-NC-SA 2.0 (30), CC BY-ND 2.0 (3), CC BY-NC-ND 2.0 (26) | 100 (open) | 71 images (open); the 29 NoDerivs images withheld |
| Face portraits | CelebA-HQ-256 | 50 | CelebA non-commercial research agreement | None (re-fetched and hash-verified) | 50 images (restricted record) |
| AI-generated | Flux Kontext Pro, author-generated | 50 | CC BY 4.0 | 50 (open) | 50 images (open) |

The perturbed inputs (the eps=0.10 images actually sent to the APIs) are withheld from both records: the perturbation generator described in §4.2 is proprietary to Stelais, and the perturbed images are worked examples of it. This does not affect the reproducibility of the corpus results. Every quantity in §5.2 and §5.4 is computed from the API outputs and the clean references, all of which are deposited; the output-vs-perturbed-input distances used descriptively in the paper are recorded per call in results.jsonl. Three further subsets are absent from the open record. (i) The 50 CelebA-HQ inputs are not redistributable under the CelebA agreement; the manifest records each one's Hugging Face source identifier (dataset Rahafbk/celebhq_train_test) and SHA-256 hash, so they can be re-fetched from the public mirror and verified byte-for-byte with the bundled verifier. (ii) Face-domain API outputs (888 files) are in the restricted-access record for non-commercial research use on request. (iii) API outputs derived from the 29 NoDerivs-licensed COCO photographs (522 files) are withheld; their metric rows remain in results.jsonl. Metadata, metrics, figures, and author-generated images are released under CC BY 4.0; outputs derived from a COCO photograph inherit that photograph's Creative Commons terms. Analysis code is available from the author on reasonable request; the perturbation generator is not available.

**Table S7.** Deposited files. Every file inside each archive is listed with its SHA-256 digest in that record's SHA256SUMS; the two large archives are deposited in 60 MB parts because a single multi-gigabyte upload could not be completed reliably.

| Record | File | Contents | Items | Size |
|---|---|---|---|---|
| open | `open_manifest.jsonl` | One row per input image: id, domain, source, SHA-256, dimensions, license, redistribution flags | 200 | 0.2 MB |
| open | `open_results.jsonl` | One row per API call: image, AI, repeat, latency, DINOv2 patch and SSIM distances, mode label; 12 rows carry the provider error string | 3,600 | 1.3 MB |
| open | `open_summary.tar` | Bootstrap CIs, domain effect, per-image consistency, per-AI × domain summary, separation figure | 5 | 1 MB |

| Record | File | Contents | Items | Size |
|---|---|---|---:|---:|
| open | `open_inputs.tar` | 150 clean inputs (100 COCO, 50 Flux) | 150 | 85 MB |
| open | `open_outputs.tar`, deposited as 49 parts `.part-000`….`part-048` with `open_outputs.tar.PARTS.sha256` and `reassemble_parts.sh` | API outputs, `<image_id>__<ai>__rep<k>.png`; reassemble with `cat open_outputs.tar.part-* > open_outputs.tar` | 2,178 | 2.9 GB |
| open | `open_coco_licenses.json`, `open_LICENSE.md`, `open_README.md`, `open_SHA256SUMS`, `open_verify_sha256.py` | Per-image COCO license table, licensing statement, documentation, digests, verifier | 5 | 0.3 MB |
| restricted | `restricted_outputs.tar`, deposited as 19 parts with its `.PARTS.sha256` and `reassemble_parts.sh` | Face-domain API outputs; reassemble as above | 888 | 1.1 GB |
| restricted | `restricted_README.md`, `restricted_SHA256SUMS` | Access conditions, digests | 2 | 0.1 MB |